\documentclass[10pt,twocolumn,letterpaper]{article}

\usepackage{cvpr}
\usepackage{times}
\usepackage{epsfig}
\usepackage{graphicx}
\usepackage{amsmath}
\usepackage{amssymb}
\usepackage{pifont}
\usepackage{bm}
\usepackage{booktabs}
\usepackage{multirow}
\usepackage{lipsum}
\DeclareMathOperator*{\fusion}{B}

\DeclareMathOperator*{\murelcell}{MurelCell}

\usepackage[pagebackref=true,breaklinks=true,letterpaper=true,colorlinks,bookmarks=false]{hyperref}

\cvprfinalcopy 

\def\cvprPaperID{846} 

\ifcvprfinal\fi
\begin{document}
\newcommand{\cmark}{\ding{51}}%
\newcommand{\xmark}{\ding{55}}%

\newcommand{\MCc}[1]{\textcolor{blue}{[Matt: #1]}}
\newcommand{\MC}[1]{{\color{blue}#1}}
\newcommand{\HB}[1]{{\color{red}#1}}

\title{MUREL: Multimodal Relational Reasoning for Visual Question Answering}

\author{Remi Cadene $^{1}$\thanks{Equal contribution}
\qquad 
Hedi Ben-younes $^{1,2}$\footnotemark[1] 
\qquad  
Matthieu Cord $^1$  
\qquad 
Nicolas Thome $^3$ 
\\ $^1$ Sorbonne Universit\'e, CNRS, LIP6, 4 place Jussieu, 75005 Paris
\\ $^2$ Heuritech, 248 rue du Faubourg Saint-Antoine, 75012 Paris 
\\ $^3$ Conservatoire National des Arts et M\'etiers, 75003 Paris
\\  {\tt\small remi.cadene@lip6.fr,
hedi.ben-younes@lip6.fr,
matthieu.cord@lip6.fr, nicolas.thome@cnam.fr}
}

\maketitle

\begin{abstract}

Multimodal attentional networks are currently state-of-the-art models for Visual Question Answering (VQA) tasks involving real images. Although attention allows to focus on the visual content relevant to the question, this simple mechanism is arguably insufficient to model complex reasoning features required for VQA or other high-level tasks. 

In this paper, we propose MuRel, a multimodal relational network which is learned end-to-end to reason over real images. Our first contribution is the introduction of the MuRel cell, an atomic reasoning primitive representing interactions between question and image regions by a rich vectorial representation, and modeling region relations with pairwise combinations.
Secondly, we incorporate the cell into a full MuRel network, which progressively refines visual and question interactions, and can be leveraged to define visualization schemes finer than mere attention maps. 

We validate the relevance of our approach with various ablation studies, and show its superiority to attention-based methods on three datasets: VQA 2.0, VQA-CP v2 and TDIUC. Our final MuRel network is competitive to or outperforms state-of-the-art results in this challenging context.

Our code is available: \href{https://github.com/Cadene/murel.bootstrap.pytorch}{\texttt{github.com/Cadene/\linebreak murel.bootstrap.pytorch}}
\end{abstract}

\section{Introduction}

Since the success of Convolutional Neural Networks (ConvNets) at the ILSVRC 2012 challenge \cite{NIPS2012_4824}, Deep Learning has become the baseline approach for any computer vision problem. 
Beyond their outstanding performances for perception tasks, \textit{e.g.} classification or detection, deep ConvNets have also been successfully used for new artificial intelligence tasks like Visual Question Answering (VQA) \cite{VQA,VQA2_Goyal_2017_CVPR,Kafle_2017_ICCV}. VQA requires a high level understanding of images and questions, and is often considered to be a good proxy for visual reasoning.
However, it is not straightforward to use ConvNets in a context where a high level of reasoning is required. The question of leveraging the perception power of deep CNNs for reasoning tasks is crucial if we want to go further in visual scene understanding \cite{johnson2016clevr, chen18iterative}.

\begin{figure}
    \centering
    \includegraphics[width=1.0\linewidth]{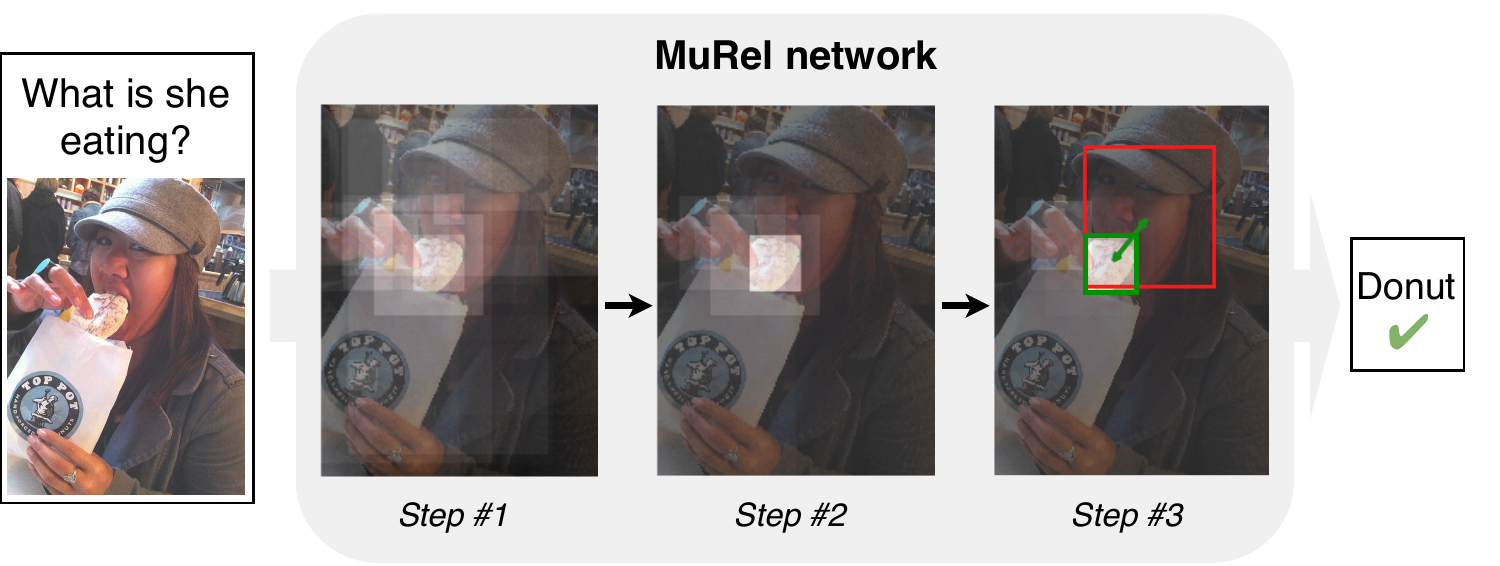}
    \caption{\label{fig:big_picture} \textbf{Visualization of the MuRel approach.}
    Our MuRel network for VQA is an iterative process based on a rich vectorial representation between the question and visual information explicitly modeling pairwise region relations. MuRel is thus able to express complex analysis primitives beyond attention maps: here the two regions corresponding to the head and the donuts are selected based on their visual cues and semantic relations to properly answer the question "what is she eating?" \vspace{-0.4cm}}
\end{figure}

It is also not trivial to define nor evaluate a model's capacity to reason about the visual modality in VQA. To fill this need, synthetic datasets have been released, \textit{e.g.} CLEVR~\cite{johnson2016clevr},  which specific structure controls the exact reasoning primitives required to give the answer \cite{johnson2017inferring, hu2017learning, Mascharka_2018_CVPR}.
However, methods that tackle the VQA problem on real data struggle to integrate this explicit reasoning procedure. Instead, state-of-the-art methods often rely on the much simpler attentional framework \cite{FukuiPYRDR16, benyounescadene2017mutan, Kim2017, chen2017sva}. Despite its effectiveness, this mechanism restricts visual reasoning to a soft selection of regions that are relevant to answer the question.
This arguably limits the modeling power of such models to bridge the gap between the perceptual strengths of ConvNets and the high-level reasoning demand for VQA.

In this paper, we propose MuRel, a multimodal relational network that goes one step further towards reasoning about questions and images.
~Our first contribution is to introduce the MuRel cell, an atomic reasoning primitive enabling to represent rich interactions between question and image regions. It is based on a vectorial representation that explicitly models relations between regions. 
Our second contribution is to embed this MuRel cell into an iterative reasoning process, which progressively refines the internal network representation to answer the question. 
The rationale of MuRel is illustrated in Figure~\ref{fig:big_picture}: for the question "what is she eating", our model focuses on two main regions (the head and the donut) with important visual cues and semantic relations between them to provide the correct answer ("donut"). 
The visual reasoning of our MuRel system is formed by this multi-step relational module that discards useless information to focus on the relevant regions.

In the experiments, we show additional results for explaining the behaviour of MuRel. We also provide various ablative studies to validate the relevance of the  MuRel cell and the iterative reasoning process, and show that MuRel is highly competitive or even outperforms state-of-the-art results on three of the most common VQA datasets: the VQA 2.0 dataset \cite{VQA2_Goyal_2017_CVPR}, VQA-CP v2 \cite{vqa-cp} and TDIUC \cite{Kafle_2017_ICCV}.

\section{Related work and contributions}

Recently, the deep learning community started to tackle complex visual reasoning problems such as relationship detection \cite{VRD_Lu_2016_ECCV}, object recognition \cite{chen18iterative}, abstract reasoning \cite{SantoroHBML18} or visual causality \cite{LopNisChiSchBot17}, while more theoretical work attempt to formalize relational reasoning \cite{DBLP:journals/corr/abs-1806-01261}.

But the most popular image reasoning task is certainly Visual Question Answering (VQA), which has been a hot research topic for the last five years \cite{malinowski2014nips, VQA, VQA2_Goyal_2017_CVPR, Kafle_2017_ICCV}. 
Since the seminal work of \cite{malinowski2014nips}, different sub-problems have been identified for the resolution of VQA. In particular, explicit reasoning techniques have been developed relying on the synthetic CLEVR dataset \cite{johnson2016clevr}. Meanwhile, real-data VQA systems are the test bed for more practical approaches based on high quality visual representations or multimodal fusion schemes.

\paragraph{Visual reasoning}
The research efforts towards VQA models that are able to reason about a visual scene is mainly conducted using the CLEVR dataset \cite{johnson2016clevr}.
This artificial dataset provides questions that require spatial and relational reasoning on simple images coming from a visual world with low variability. 
An important line of work attempts to solve this task through explicit reasoning. In such methods \cite{johnson2017inferring, hu2017learning, Mascharka_2018_CVPR}, a neural network reads the question and generates a program, corresponding to a graph of elementary neural operations that process the image. 
However, there are two major downsides to these techniques. First, their performance strongly depends on whether or not program annotations are used to learn the program generator; 
and second, they can be matched or surpassed by simpler models that implicitly learn to reason without requiring program annotation.
In particular, FiLM \cite{perez2018film} modulates the visual feature map with an affine transformation whose parameters depend on the question. 
In more recent work, the MAC network \cite{arad2018compositional} draws inspiration from the Model-View-Controller paradigm to design the trainable MAC cell on which the network iterates.
Finally, in \cite{SantoroRBMPBL17}, they reason over all the possible pairs of objects in the picture, thus introducing relationship modeling in visual question answering.

\paragraph{VQA on real data}
An important part of the research in VQA is focused on designing functions that can represent high-level correlations between two vector spaces.
Among these  multimodal fusion algorithms, the most effective ones use second order (or higher \cite{yu2018beyond}) interactions, made tractable through sketching methods
\cite{FukuiPYRDR16}, or with more success using the tensor decomposition framework \cite{Kim2017,benyounescadene2017mutan,yu2017mfb}.

This line of work is often considered orthogonal to visual reasoning contributions. 
In a setup involving real data, complex methods such as explicit or relational reasoning are much more challenging to implement than with artificial images and questions.
This is certainly why the most widely used reasoning framework involves soft attention mechanisms \cite{bahdanau+al-2014-nmt, Xu:2015:SAT:3045118.3045336}. Given a question, these models assign an importance score to each region, and use them to weight-sum pool the visual representations. Multiple attention maps (also called \textit{glimpses}) can even be computed in parallel \cite{Kim2017,benyounescadene2017mutan,yu2017mfb,yu2018beyond} or sequentially \cite{YangHGDS16}. More complex attention strategies have been explored, such as the Structured Attention \cite{chen2017sva}, where a locally-connected graphical structure is considered to infer the region saliency scores. 
\cite{zhang2018learning} also leverages a graphical structure between regions to address weaknesses of the soft-attention mechanism, improving the VQA model's ability to count. 
In \cite{learningconditionedgraph}, the image representation is computed using pairwise semantic attention and spatial graph convolutions.
The soft attention framework is questioned in \cite{Malinowski_2018_ECCV}, where regions are hardly selected based on the norm of their feature. Finally, recent work of \cite{Kim2018} simultaneously attends over regions and word tokens through a bilinear attention network.

Importantly, the type of visual features used to feed the VQA system has an large impact on performance.
While early work have been using fixed-grid representation given by a fully-convolutional network (such as ResNet-152 \cite{DBLP:conf/cvpr/HeZRS16}), performance can be improved using predictions from an object detector \cite{Anderson_2018_CVPR}.
Recently, a crucial component in the VQA Challenge 2018 winning entry was the mix of multiple types of visual features \cite{pythia18arxiv}.

\paragraph{MuRel contributions}
In this work, we move away from the classical attention framework \cite{Kim2017, FukuiPYRDR16, benyounescadene2017mutan, yu2018beyond} widely used in real-data VQA systems. 
Instead, we use a vectorial representation, more expressive than scalar attention maps, to model the semantic interaction between each region's visual content and the question.
In addition, we include a notion of spatial and semantic context in the representations by representing pairs of image regions through interactions between their visual embeddings and spatial coordinates.
Differently than the approach followed in~\cite{learningconditionedgraph} where a locally connected graph structure is built, we use the relations between all possible pairs of regions.

Our MuRel network embodies an iterative process with inspiration from works driven by the synthetic reasoning CLEVR dataset, \textit{e.g.}, MAC \cite{arad2018compositional} or FiLM \cite{perez2018film}, which we adapt to the real data VQA purpose. In particular, we improve the interactions between image regions and questions by using richer bilinear fusion models and by explicitly incorporating relations between regions.
\section{MuRel approach}
Our VQA approach is depicted in Figure~\ref{fig:murel_archi}. Given an image $v \in \mathcal{I}$ and a question $q \in \mathcal{Q}$ about this image, we want to predict an answer $\hat{a} \in \mathcal{A}$ that matches the ground truth answer $a^\star$. As very common in VQA, the prediction $\hat{a}$ is given by classification scores:
\begin{equation}
    \hat{a} = \arg\!\max_{a \in \mathcal{A}} p_\theta(a | v,q)
\end{equation}

where $p_\theta$ is our trainable model. In our system, the image is represented by a set of vectors $\{\bm{v}_i \}_{i\in[1,N]}$, where each $\bm{v}_i \in \mathbb{R}^{d_v}$ corresponds to an object detected in the picture. We also use the spatial coordinates of each region $\bm{b}_i~=~[x, y, w, h]$, where $(x,y)$ are the coordinates of the top-left point of the box, and $h$ and $w$ correspond to the height and the width of the box. Note that $x$ and $w$ (respectively $y$ and $h$) are normalized by the width (resp. height) of the image.
For the question, we use a gated recurrent unit network to provide a sentence embedding $\bm{q} \in \mathbb{R}^{d_q}$. 

In Section~\ref{sub:MuRelcell}, we present the MuRel cell, a neural module that learns to perform elementary reasoning operations by blending question information into the set of spatially-grounded visual representations. Next, in Section~\ref{sub:murelnet}, we leverage the power of this cell using the MuRel network, a VQA architecture that iterates through a MuRel cell to reason about the scene with respect to a question.

\subsection{MuRel cell}
\label{sub:MuRelcell}

\begin{figure}
    \centering
    \includegraphics[width=\linewidth]{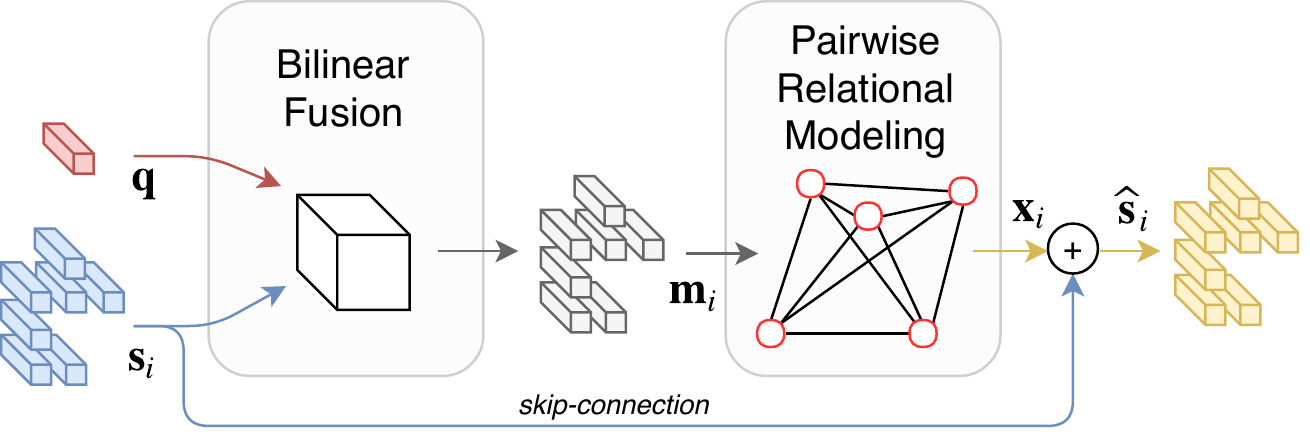}
    \caption{\label{fig:murel_cell} \textbf{MuRel cell.}
    In the MuRel cell, the bilinear fusion represents rich and fine-grained interactions between question and region vectors $\bm{q}$ and $\bm{s}_i$. All the resulting multimodal vectors $\bm{m}_i$ pass through a pairwise modeling block to provide a context-aware embedding $\bm{x}_i$ per region. 
    The cell's output $\hat{\bm{s}}_i$ is finally computed as a sum between ${\bm{s}_i}$ and $\bm{x}_i$, acting as residual function of ${\bm{s}_i}$.}
\end{figure}

The MuRel cell takes as input a bag of $N$ visual features $\bm{s}_i \in \mathbb{R}^{d_v}$, along with their bounding box coordinates $\bm{b}_i$.
As shown in Figure~\ref{fig:murel_cell}, it is a residual function consisting of two modules. First, an efficient bilinear fusion module merges question and region feature vectors to provide a local multimodal embedding. This fusion is directly followed by a pairwise modeling component, designed to update each multimodal representation with respect to its own spatial and visual context.

\paragraph{Multimodal fusion}
We want to include question information within each visual representation $\bm{s}_i$. 
Multiple multimodal fusion strategies have been recently proposed \cite{Kim2017, FukuiPYRDR16, benyounescadene2017mutan, yu2017mfb, yu2018beyond} to model the relevant interactions between two modalities. One of the most efficient technique is the one proposed by \cite{benyounescadene2017mutan}, based on the Tucker decomposition of third-order tensors. 
This bilinear fusion model learns to focus on the relevant correlations between input dimensions. It models rich and fine-grained multimodal interactions, while keeping a relatively low number of parameters. 
Each input vector $\bm{s}_i$ is fused with the question embedding $\bm{q}$ using the same bilinear fusion:
\begin{equation}
    \bm{m}_i = \fusion \left( \bm{s}_i, \bm{q}; \Theta\right)
\end{equation}
where $\Theta$ are the trainable parameters of the fusion module. Each dimension $m$ of $\bm{m}_i$ can be written as a bilinear function in the form $\sum_{s,q} w^{s,q,m}\bm{s}_i^s \bm{q}^q$. Thanks to the Tucker decomposition, the tensor $\{w_{s,q,m}\}$ is factorized into the list of parameters $\Theta$. We set the number of dimensions in $\bm{m}_i$ to $d_v$ to facilitate the use of residual connections throughout our architecture.

In classical attention models, the fusion between image region and question features $\bm{s}$ and $\bm{q}$ only learns to encode whether a region is relevant. 
In the MuRel cell, the local multimodal information is represented within a richer vectorial form $\bm{m}_i$ which can encode more complex correlations between both modalities. This allows to store more specific information about what precise characteristic of a particular region is important in a given textual context.

\paragraph{Pairwise interactions}
To answer certain types of question, it can be necessary to reason over multiple object that interact together. More generally, we want each representation to be aware of the spatial and semantic context around it.
Given that our features are structured as a bag of localized vectors \cite{Anderson_2018_CVPR}, modeling the visual context of each region is not straightforward. Similarly to the recent work of \cite{learningconditionedgraph}, we opt for a pairwise relationship modeling where each region receives a message based on its relations to its neighbours. 
In their work, a region's neighbours correspond to the $K$ most similar regions, whereas in the MuRel cell the neighbourhood is composed of every region in the image.
Besides, instead of using scalar pairwise attention and graph convolutions with Gaussian kernels as they do, we merge spatial and semantic representations to build relationship vectors. 
In particular, we compute a context vector $\check{\bm{e}}_i$ for every region. It consists in an aggregation of all the pairwise links $\bm{r}_{i,j}$ coming into $i$. 
We define it as $\check{\bm{e}}_i = \max_j \bm{r}_{i,j}$, where $\bm{r}_{i,j}$ is a vector containing information about the content of both regions, but also about their relative spatial positioning.
We use the $\max$ operator in the aggregation function to reduce the noise that can be induced by average or sum poolings, which oblige all the regions to interact with each other.
To encode the relationship vector, we use the following formulation:

\begin{equation}
    \bm{r}_{i,j} = \fusion \left(\bm{b}_i, \bm{b}_j; \Theta_b \right) + \fusion \left( \bm{m}_i, \bm{m}_j; \Theta_m \right)
\end{equation}

Through the $\fusion(.,.;\Theta_b)$ operator, the cell is free to learn spatial concepts such as \textit{on top of, left, right, etc.} In parallel, $\fusion(.,.;\Theta_s)$ encodes correlations between multimodal vectors $(\bm{s}_i, \bm{s}_j)$, corresponding to semantic visual concepts conditionned on the question representation. By summing up both spatial and semantic fusions, the network can learn high-level relational concepts such as \textit{wear, hold, etc.}

The context representation $\check{\bm{e}}_i$ that contains an aggregation of the messages $\bm{r}_{i,j}$ provided by its neighbours updates the multimodal vector $\bm{m}_i$ in an additive manner:

\begin{equation}
\label{eq:pairwise}
    \bm{x}_i = \bm{m}_i + \check{\bm{e}}_i
\end{equation}
This formulation of the pairwise modelling is actually closer to the Graph Networks \cite{DBLP:journals/corr/abs-1806-01261}, where the notion of relational inductive biases is formalized.

Finally, the MuREL cell's output is computed as a residual function of its input, to avoid the vanishing gradient problem. Each visual feature $\bm{s}_i$ is updated as: $\hat{\bm{s}}_i = \bm{s}_i + \bm{x}_i$.

The chain of operations that updates the set of localized region embeddings $\{\bm{s}_i\}_{i \in [1,N]}$ using the multimodal fusion with $\bm{q}$ and the pairwise modeling operator is noted:
\begin{equation}
    \{\hat{\bm{s}}_i\} = \murelcell \left( \{\bm{s}_i\}; \{\bm{b}_i\}, \bm{q} \right)
\end{equation}

\subsection{MuRel network}
\label{sub:murelnet}
\begin{figure*}
    \centering
    \includegraphics[width=\textwidth]{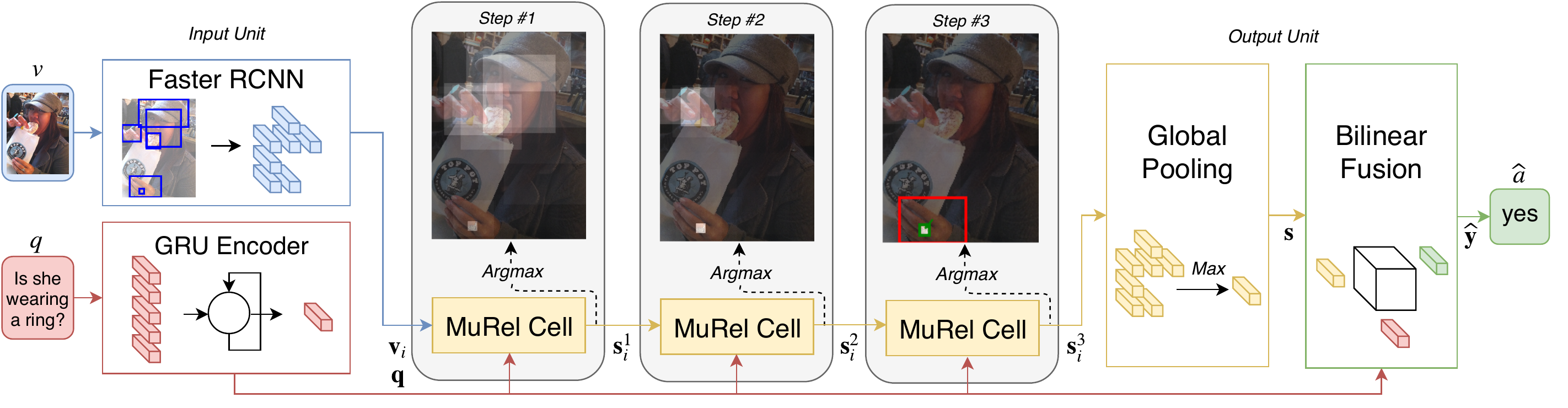}
    \caption{\label{fig:murel_archi} \textbf{MuRel network.} The MuRel network merges the question embedding $\bm{q}$ into spatially-grounded visual representations $\{\bm{v}_i\}$ by iterating through a single MuRel cell. This module takes as input a set of localized vectors $\{\bm{s}_i\}$ and updates their representation using a multimodal fusion component. Moreover, it models all the possible pairwise relations between regions by combining spatial and semantic information. To construct the importance map at step $t$, we count the number of time each region provides the maximal value of $\max_i \{\bm{s}_i^t\}$ (over the 2048 dimensions).}
\end{figure*}

Mimicking a simple form of progressive reasoning, our model leverages the power of bilinear fusions to iteratively merge visual information into context-aware visual embeddings.
As we can see in Figure~\ref{fig:murel_archi}, a MuRel cell iteratively updates the region state vectors $\{\bm{s}_i\}$, each time refining the representations with contextual and question information. More specifically, for each step $t=1..T$ where $T$ is the total number of steps, a MuRel cell processes and updates the state vectors following Equation~\eqref{eq:iteration}:

\begin{equation}
\label{eq:iteration}
    \{\bm{s}_i^{t}\} = \murelcell \left( \{\bm{s}_i^{t-1}\}; \{\bm{b}_i\}, \bm{q} \right)
\end{equation}

The state vectors are initialized with the features outputted by the object detector; for each region $i$,  $\bm{s}_i^0 = \bm{v}_i$. 

The MuRel network represents each region regarding the question, but also using its own visual context. This representation is done iteratively, through multiple steps of a MuRel cell. 
The residual nature of this module makes it possible to align multiple cells without being subject to gradient vanishing. 
Moreover, the weights of our model are shared across the cells, which enables compact parametrization and good generalization.

At step $t=T$, the representations $\{ \bm{s}_i^T \}$ are aggregated with a global max pooling operation to provide a single vector $\bm{s} \in \mathbb{R}^{d_v}$. This scene representation contains information about the objects, the spatial and semantic relations between them, with respect to a particular question.

The scene representation $\bm{s}$ is merged with the question embedding $\bm{q}$ to compute a score for every possible answer $\hat{\bm{y}} = \fusion \left( \bm{s}, \bm{q}; \Theta_{y} \right)$. Finally, $\hat{a}$ is the answer with maximum score in $\hat{\bm{y}}$.

\paragraph{Visualizing MuRel network}
Our model can also be leveraged to define visualization schemes finer than mere attention maps. 
Especially, we can highlight important relations between image regions for answering a specific question.
At the end of the MuRel network, the visual features $\{\bm{s}_i^T\}$ are aggregated using a $\max$ operation, yielding a $d_v-$dimensional vector $\bm{s}$. 
Thus, we can compute a \textit{contribution map} by measuring to what extent each region contributes to the final vector. 
To do so, we compute the point-wise $\bm{c} = \arg\!\max_i  \{\bm{s}_i^T\}  \in [1, N]^{d_v}$, and measure the occurrence frequency of each region in this vector $\bm{c}$. This provides a value for each region that estimates its contribution to the final vector.
Interestingly, this process can be done after each cell, and not exclusively at the last one. Intuitively, it measures what the contribution map would have been if the iterative process had stopped at this point.
As we can see in Figures~\ref{fig:big_picture},\ref{fig:murel_archi},\ref{fig:quali}, these relevance scores match human intuition and can be used to explain the model's decision, even if the network has not been trained with any selection mechanism.

Similarly, we are able to visualize the pairwise relationships involved in the prediction of the MuRel cell. The first step is to find $i^\star$, which is the region that is the most impacted by the pairwise modeling. It is the region such that $\|\frac{\check{\bm{e}}_i}{\bm{x}_i} \|_2$ is maximal (cf. Equation~\eqref{eq:pairwise}). This bounding box is shown in green in all our visualizations. We then measure the contribution of every other region to $i^\star$ using the occurrence frequencies in $\arg\!\max_j \bm{r}_{i,j}$. We show in red the regions whose contribution to $i^\star$ is above a certain threshold (0.2 in our visualizations). If there is no such region, the green box is not shown.

\paragraph{Connection to previous work}
We can draw a comparison between our MuRel network and the FiLM network proposed in \cite{perez2018film}. 
Beyond the fact that their model is built for the synthetic CLEVR dataset \cite{johnson2016clevr} and ours processes real data, some connections can be found between both models.
In their work, the image passes through multiple residual cells, whereas we only have one cell through which we iterate. In FiLM, the multimodal interaction is modeled with a feature-wise affine modulation, while we use a bilinear fusion strategy \cite{benyounescadene2017mutan} which seems better suited to real world data. 
Finally, both MuRel and FiLM leverage the spatial structure of the image representation to model the relations between regions.
In FiLM, the image is represented with a fully-convolutional network which outputs a feature map disposed in a fixed spatial grid. With this structure on image features, the relations between regions are modeled with a $3 \times 3$ convolution inside each residual block. Thus, the representation of each region depends on its neighbours in the locally-connected graph induced by the fixed grid structure. 
In our MuRel network, the image is represented as a set of localized features. This makes the relational modeling non trivial. As we want to model relations between regions that are potentially far apart, we consider that the set of regions forms a complete graph, where each region is connected to all the others. 
\section{Experiments}

\subsection{Experimental setup}

\textbf{Datasets}: We validate the benefits of the MuRel cell and the MuRel network on three recent datasets. 
VQA 2.0 \cite{VQA2_Goyal_2017_CVPR} is the most used dataset. It comes with a training set, a validation set and an online testing set. We provide a fine grained analysis on the validation set, while we compare MuRel to the state-of-the-art models on the testing set.
Then, we use VQA Changing Priors v2 \cite{agrawal2018don} to demonstrate the generalization capacity of MuRel. VQA-CP v2 uses the same data as in VQA 2.0, but proposes different distribution of answers per question between training and validation splits.
Finally, we use the TDIUC dataset \cite{Kafle_2017_ICCV} to construct a more detailed analysis of our model's performance on 12 well-defined types of question. TDIUC is currently the biggest dataset for visual question answering.

\textbf{Hyper-parameters}: We use standard features extraction, preprocessings and loss function \cite{FukuiPYRDR16}. 
We use the recent Bottom-up features provided by \cite{Anderson_2018_CVPR} to represent our image as a set of 36 localized regions. 
For the question embedding, we use the pretrained Skip-thought encoder from \cite{Kiros2015}. 
Inspired by recent works, we use Adam as optimizer \cite{KingmaB14} with a learning scheduler \cite{pythia18arxiv}. 
More details about the experimental setup are given in appendix.

\subsection{Model validation}

We compare MuRel against models trained on the same Bottom-up features \cite{Anderson_2018_CVPR} which are required to reach the best performances.

\begin{table}[]
    \centering
    \begin{tabular}{*5{c}}
    \toprule
        Model & VQA 2.0 & VQA CP v2 & TDIUC \\
    \midrule
        Attention baseline & 63.44 & 38.04 & 86.96 \\ 
        MuRel              & \textbf{65.14} & \textbf{39.54} & \textbf{88.20 }\\ 
    \bottomrule
    \end{tabular}
    \vspace{0.1cm}
    \caption{\textbf{Comparing MuRel to Attention.} Comparison of the MuRel strategy against a strong Attention-based model on the VQA 2.0 \textit{val}, VQA-CP v2 and TDIUC datasets. Both models have an equivalent number of parameters ($\sim$60 million) and are trained on the same features following the same experimental setup.}
    \label{tab:model_validation}
\end{table}

\begin{table}[]
    \centering
    \begin{tabular}{*2{c}@{\hskip 0.15in}|@{\hskip 0.15in}*3{c}}
    \toprule
        Pairwise & Iter. & VQA 2.0 & VQA CP v2 & TDIUC \\
    \midrule
        \xmark & \xmark & 64.13 & 38.88 & 87.50 \\ 
        \cmark & \xmark & 64.57 & 39.12 & 87.86 \\ 
        \xmark & \cmark & 64.72 & 39.37 & 87.92 \\ 
        \cmark & \cmark & \textbf{65.14} & \textbf{39.54} & \textbf{88.20 }\\ 
    \bottomrule
    \end{tabular}
    \vspace{0.1cm}
    \caption{\textbf{Ablation study of MuRel}. Experimental validation of the pairwise module and the iterative processing on the VQA 2.0 \textit{val}, VQA-CP v2 and TDIUC datasets.}
    \label{tab:ablation}
\end{table}

\paragraph{Comparison to Attention-based model}
In Table~\ref{tab:model_validation}, we compare MuRel against a strong attentional model based on bilinear fusions \cite{benyounescadene2017mutan}, which encompasses a  multi-glimpses attentional process \cite{FukuiPYRDR16}. The goal of this experiments is to compare our approach with strong baselines for real VQA in controlled conditions. In addition to using the same bottom-up features, which are crucial for fair comparisons, we also dimension the attention-based baseline to have an equivalent amount of learned parameters than MuRel ($\sim$60 millions including those from the GRU encoder). Also, we train it following the same experimental setup to insure competitiveness.
MuRel reaches a higher accuracy on the three datasets. We report a significant gain of +1.70 on VQA 2.0 and +1.50 on VQA CP v2. 
Not only these results validate the ability of MuRel to better model interactions between the question and the image, but also to generalize when the distribution of the answers per question are completely different between the training and validation set as in VQA CP v2.
A gain of +1.24 on TDIUC demonstrates the richer modeling capacity of MuRel in a fine-grained context of 12 well delimited question types. 

\paragraph{Ablation study} In Table~\ref{tab:ablation}, we compare three ablated instances of MuRel to its complete form.
First, we validate the benefits of the pairwise module. Adding it to a vanilla MuRel without iterative process leads to higher accuracy on every datasets. In fact, between line 1 and 2, we report a gain of +0.44 on VQA 2.0, +0.24 on VQA CP v2 and +0.36 on TDIUC.
Secondly, we validate the interest of the iterative process. Between line 1 et 3, we report a gain of +0.59 on VQA 2.0, +0.49 on VQA CP v2 and +0.42 on TDIUC.
Notably, this modification does not add any parameters, because we iterate over a single MuRel cell. 
Unsharing the weights by using a different MuRel cell for each step gives similar results.
Finally, the pairwise module and the iterative process are added to create the complete MuRel network. This instance (in line 4) reaches the highest accuracy on the three datasets. Interestingly, the gains provided by the combination of the two methods are sometimes larger than those of each one separately. For instance, we report a gain of +1.01 on VQA 2.0 between line 1 and 4. This attests to the complementary of the two modules.

\begin{figure}
    \centering
    \includegraphics[width=\linewidth]{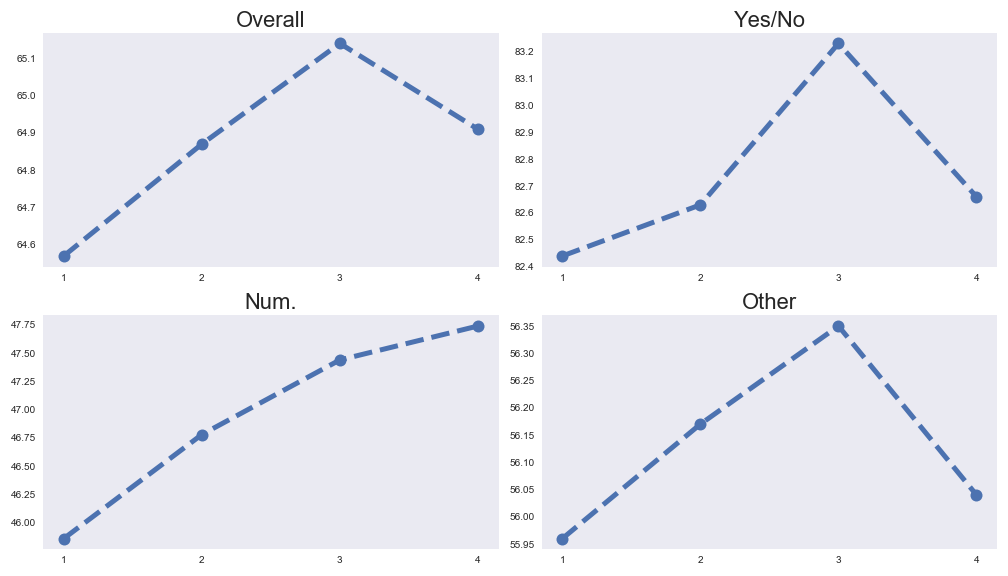}
    \caption{\label{fig:impact_nb_cell} \textbf{Number of iterations}. Impact of the number of steps in the iterative process on the different question types of VQA 2.0 \textit{val}.}
\end{figure}

\paragraph{Number of reasoning steps} In Figure~\ref{fig:impact_nb_cell},
we perform an analysis of the iterative process. 
We train four different MuRel networks on the VQA 2.0 \textit{train} split, each with a different number of iterations over the MuRel cell. Performance is reported on \textit{val} split.
Networks with two and three steps respectively provides a gain of +0.30 and +0.57 in overall accuracy on VQA 2.0 over the network with a single step. 
An interesting aspect of the iterative process of MuRel is that the four networks have exactly the same amount of parameters, but the accuracy significantly varies with respect to the number of steps.
While the accuracy for the answer type involving numbers keeps increasing, we report a decrease in overall accuracy at four reasoning steps. 
Counting is a challenging task: not only does the model need to detect every occurrence of the desired object, but also the representation computed after the final aggregation must keep the information of the number of detected instances. The complexity of this question may require deeper relational modeling, and thus benefit from a higher number of iterations over the MuRel cell.

\subsection{State of the art comparison}

\paragraph{VQA 2.0}
In Table~\ref{tab:vqa2_sota}, we compare MuRel to the most recent contributions on the VQA 2.0 dataset. For fairness considerations, all the scores correspond to models trained on the VQA 2.0 \textit{train+val} split, using the Bottom-up visual features \cite{Anderson_2018_CVPR}. 
Interestingly, our model surpasses both MUTAN \cite{benyounescadene2017mutan} and MLB \cite{Kim2017}, which correspond to some of the latest development in visual attention and bilinear models. This tends to indicate that VQA models can benefit from retaining local information in mulitmodal vectors instead of scalar coefficients. 
Moreover, our model greatly improves over the recent method proposed in \cite{learningconditionedgraph} where the regions are structured using pairwise attention scores, which are leveraged through spatial graph convolutions. This shows the interest of our spatial-semantic pairwise modeling between all possible pairs of regions.
Finally, even though we did not extensively tune the hyperparameters of our model, our overall score on the \textit{test-dev} split is highly competitive with state-of-the-art methods. In particular, we are comparable to Pythia \cite{pythia18arxiv} who won the VQA Challenge 2018. Please note that they improve their overall scores up to 70.01\% when they include multiple types of visual features and more training data. Also, we did not report the score of 69.52\% obtained by BAN \cite{Kim2018} as they train their model on extra data from the Visual Genome dataset \cite{Krishna_2017_IJCV}.

\begin{table}
    \centering
    \begin{tabular}{*5{c}c}
        \toprule
        & \multicolumn{4}{c}{\textit{test-dev}} & \textit{test-std}\\
        Model & Yes/No & Num. & Other & All & All  \\
        \midrule
        Bottom-up & \multirow{2}{*}{81.82} & \multirow{2}{*}{44.21} & \multirow{2}{*}{56.05} & \multirow{2}{*}{65.32} & \multirow{2}{*}{65.67} \\
        \cite{Anderson_2018_CVPR} \\
        Graph Att. & \multirow{2}{*}{-} & \multirow{2}{*}{-} & \multirow{2}{*}{-} & \multirow{2}{*}{-} & \multirow{2}{*}{66.18}\\
        \cite{learningconditionedgraph}\\
        MUTAN$\dagger$ & \multirow{2}{*}{82.88} & \multirow{2}{*}{44.54} & \multirow{2}{*}{56.50} & \multirow{2}{*}{66.01} & \multirow{2}{*}{66.38}\\
         \cite{benyounescadene2017mutan}  \\
        MLB$\dagger$ & \multirow{2}{*}{83.58} & \multirow{2}{*}{44.92} & \multirow{2}{*}{56.34} & \multirow{2}{*}{66.27} & \multirow{2}{*}{66.62}\\
        \cite{Kim2017} \\
        DA-NTN & \multirow{2}{*}{84.29} & \multirow{2}{*}{47.14} & \multirow{2}{*}{57.92} & \multirow{2}{*}{67.56} & \multirow{2}{*}{67.94}\\
        \cite{Bai_2018_ECCV} \\
        Pythia & \multirow{2}{*}{-} & \multirow{2}{*}{-} & \multirow{2}{*}{-} &  \multirow{2}{*}{68.05} & \multirow{2}{*}{-} \\
        \cite{pythia18arxiv}\\
        Counter & \multirow{2}{*}{83.14} & \multirow{2}{*}{\textbf{51.62}} & \multirow{2}{*}{\textbf{58.97}} & \multirow{2}{*}{\textbf{68.09}} & \multirow{2}{*}{\textbf{68.41}}\\
         \cite{zhang2018learning} \\
        \midrule
        MuRel & \textbf{84.77} & 49.84 & 57.85 & 68.03 & \textbf{68.41} \\ 
        \bottomrule
    \end{tabular}
    \vspace{0.1cm}
    \caption{\label{tab:vqa2_sota} \textbf{State-of-the-art comparison on the VQA 2.0 dataset.} Results on \textit{test-dev} and \textit{test-std} splits. All these models were trained on the same training set (VQA 2.0 \textit{train+val}), using the Bottom-up features provided by \cite{Anderson_2018_CVPR}. No ensembling methods have been used. $\dagger$ have been trained by \cite{Bai_2018_ECCV}.}
\end{table}

\paragraph{TDIUC}
One of the core aspect of VQA models lies in their ability to address different tasks. The TDIUC dataset enables a detailed analysis of the strengths and limitations of a model by evaluating its performance on different types of question. We show in Table~\ref{tab:tdiuc_sota} a detailed comparison of recent models to our MuRel. We obtain state-of-the-art results on the Overall Accuracy and the arithmetic mean of per-type accuracies (A-MPT), and surpass by a significant margin the second best model proposed by \cite{Shi_2018_ECCV}. Interestingly, we improve over this model even though it uses a combination of Bottom-up and fixed-grid features, as well as a supervision on the question types (hence its 100\% result on the \textit{Absurd} task).
MuRel notably surpasses all previous methods on the Positional reasoning (+5.9 over MCB), Counting (+8.53 over QTA) questions. These improvements are likely due to the pairwise structure induced within the MuRel cell, which makes the answer prediction depend on the spatial and semantic relations between regions. The effectiveness of our per-region context modelling is also demonstrated by our the improvement on Scene recognition questions. For these questions, representing the image as a collection of independent objects shows lower performance than replacing each of them in its spatial and semantic context.
Interestingly, our results on the harmonic mean of per-type accuracies (H-MPT) are lower than state-of-the-art. For MuRel, this harmonic metric is significantly harmed by our low score of 21.43\% on the \textit{Utility and Affordances} task. As these questions concern the possible usages of objects present in the scene (such as \textit{Can you eat the yellow object?}), and are not directly related to the visual understanding of the scene.

\begin{table}
    \centering
    \begin{tabular}{*4{c}|c}
    \toprule 
         & RAU*  & MCB*  & QTA  &  \multirow{2}{*}{MuRel} \\
         & \cite{Noh_2016_Arxiv} & \cite{FukuiPYRDR16} & \cite{Shi_2018_ECCV} & \\
         \midrule
         Bottom-up & \xmark & \xmark & \cmark & \cmark \\ 
         \midrule
         Scene Reco. & 93.96 & 93.06 & 93.80 & \textbf{96.11}\\
         Sport Reco. & 93.47 & 92.77 & 95.55 &  \textbf{96.20}\\
         Color Attr. & 66.86 & 68.54 & 60.16 &  \textbf{74.43}\\
         Other Attr. & 56.49 & 56.72 & 54.36 &  \textbf{58.19}\\
         Activity Reco. & 51.60  & 52.35 & 60.10 &  \textbf{63.83}\\
         Pos. Reasoning & 35.26 & 35.40 & 34.71 &  \textbf{41.19}\\
         Object Reco. & 86.11 & 85.54 & 86.98 &  \textbf{89.41}\\
         Absurd & 96.08 & 84.82 & \textbf{100.00} & 99.8\\
         Util. and Afford. & 31.58 & \textbf{35.09} & 31.48 &  21.43\\
         Object Presence & 94.38 & 93.64 & 94.55 &  \textbf{95.75}\\
         Counting & 48.43 & 51.01 & 53.25 & \textbf{61.78}\\
         Sentiment & 60.09 & \textbf{66.25} & 64.38 &  60.65\\
         \midrule
         Overall (A-MPT) & 67.81 & 67.90 & 69.11 & \textbf{71.56}\\
         Overall (H-MPT)&  59.00 & \textbf{60.47} & 60.08 & 59.30\\
         \midrule 
         Overall Accuracy & 84.26 & 81.86 & 85.03 & \textbf{88.20}\\
         \bottomrule 
    \end{tabular}
    \vspace{0.1cm}
    \caption{\textbf{State-of-the-art comparison on the TDIUC dataset.} *~trained by \cite{Kafle_2017_ICCV}.}
    \label{tab:tdiuc_sota}
\end{table}

\begin{figure*}[t!]
    \centering
    \includegraphics[width=\linewidth]{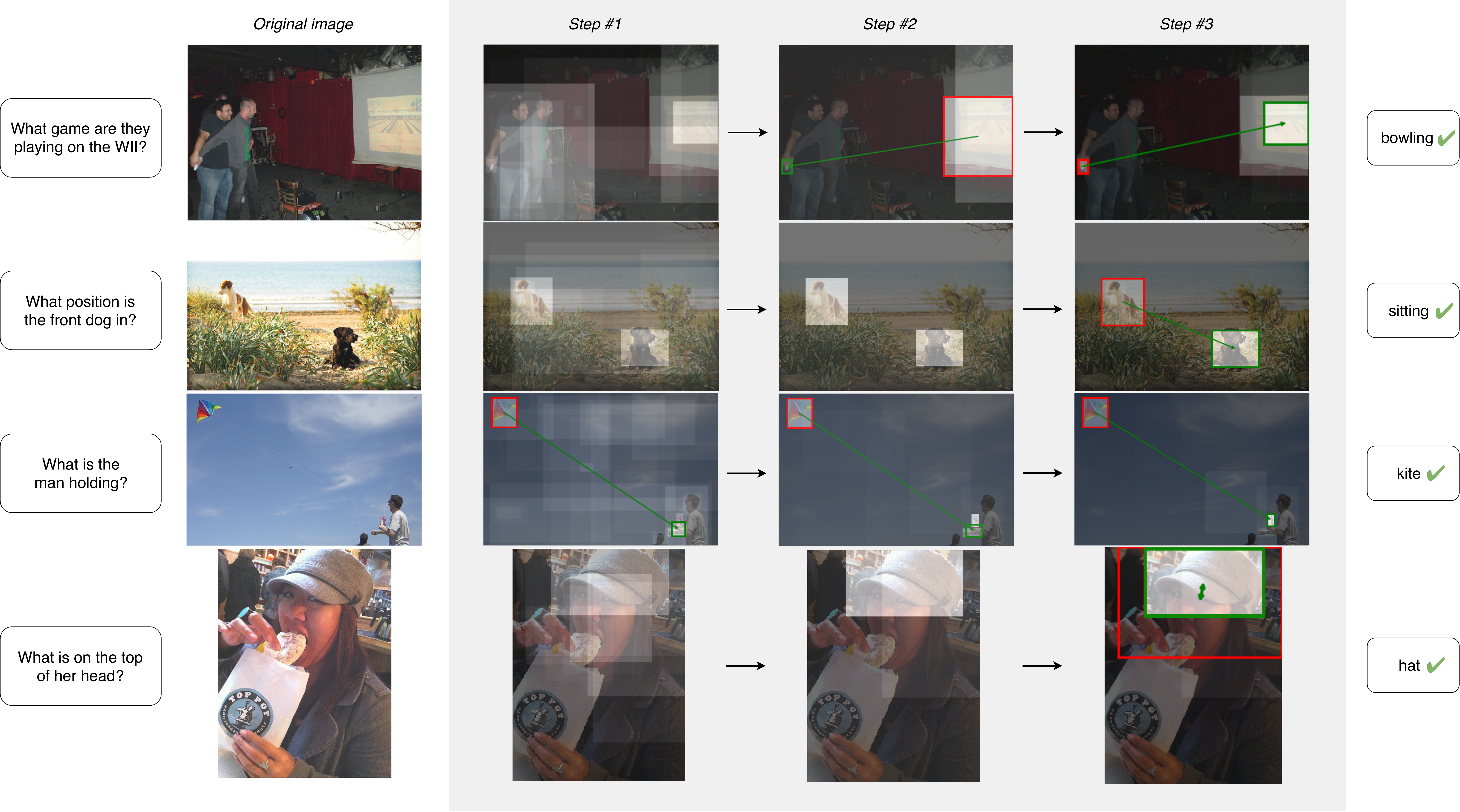}
    \caption{\label{fig:quali}\textbf{Qualitative evaluation of MuRel}. Visualization of the importance maps with colored regions related to the relational mechanism. As in Figure \ref{fig:murel_archi}, the most selected regions by the implicit attentional mechanism are shown in brighter. The green region is the most impacted by the pairwise modeling, while the red regions impact the green regions the most. These colored regions are only represented if they are greater than a certain threshold.}
\end{figure*}

\paragraph{VQA-CP v2}
This dataset has been proposed to evaluate and reduce the question-oriented bias in VQA models. In particular, the distributions of answers with respect to question types differ from \textit{train} to \textit{val} splits. 
In Table~\ref{tab:vqacp_sota}, we report the scores of two recent baselines \cite{vqa-cp, Malinowski_2018_ECCV}, on which we improve significantly. In particular, we demonstrate an important gain over GVQA \cite{vqa-cp}, whose architecture is designed to focus on Yes/No questions. However, since both methods do not use the Bottom-up features, the fairness of the comparison can be questioned. So we also train an attention model similar to \cite{benyounescadene2017mutan} using these Bottom-up region representation. 
We observe that MuRel provides a substantial gain over this strong attention baseline. Given the distribution mismatch between \textit{train} and \textit{val} splits, models that only focus on linguistic biases to answer the question are systematically penalized on their \textit{val} scores. This property of VQA-CP v2 implies that the pairwise iterative structure of MuRel is less prone to question-based overfitting than classical attention architectures. 

\begin{table}
    \centering
    \begin{tabular}{*6{c}}
        \toprule
        \multirow{2}{*}{Model} & Bottom & \multirow{2}{*}{Yes/No} & \multirow{2}{*}{Num.} & \multirow{2}{*}{Other} & \multirow{2}{*}{All}  \\
        & up \\
        \midrule
        HAN \cite{Malinowski_2018_ECCV} & \xmark & 52.25 & \textbf{13.79} & 20.33 & 28.65 \\
        GVQA \cite{vqa-cp} & \xmark & \textbf{57.99} & 13.68 & 22.14 & 31.30  \\
        Attention & \cmark & 41.56  & 12.19 & 43.29 & 38.04 \\ 
        \midrule
        MuRel & \cmark & 42.85 & 13.17 & \textbf{45.04}  & \textbf{39.54} \\ 
        \bottomrule
    \end{tabular}
    \vspace{0.1cm}
    \caption{\label{tab:vqacp_sota}\textbf{State-of-the-art comparison on the VQA-CP v2 dataset.} The Attention model was trained by us using the Bottom-up features.}
\end{table}

\subsection{Qualitative results}
In Figure~\ref{fig:quali} we illustrate the behaviour of a MuRel network with three shared cells.
Iterations through the MuRel cell tend to gradually discard regions, keeping only the most relevant ones. 
As explained in Section~\ref{sub:murelnet}, the regions that are most involved in the pairwise modeling process are shown in green and red.
Both region contributions and pairwise links match human intuition. In the first row, the most relevant relations according to our model are between the player's hand, containing the WII controller, and the screen, which explains the prediction \textit{bowling}. In the third row, the model answers \textit{kite} using the relation between the man's hand and the kite he is holding. Finally, in the last row, our model is able to address a third question on the same image than in Figure \ref{fig:big_picture} and \ref{fig:murel_archi}. Here, the relation between the head of the woman and her hat is used to provide the right answer. As VQA models are often subject to linguistic bias \cite{VQA2_Goyal_2017_CVPR,vqa-cp}, this type of visualization shows that the MuRel network actually relies on the visual information to answer questions.

\section{Conclusion}

In this paper, we introduced MuRel, a multimodal relational network for Visual Question Answering task. Our system is based on rich representations of visual image regions that are progressively merged with the question  representation.  We also included region relations with pairwise combinations in our fusion, and the whole system can be leveraged to define visualization schemes helping to interpret the decision process of MuRel.

We validated our approach on three challenging datasets: VQA 2.0, VQA-CP v2 and TDIUC.
We exhibited various ablation studies, clearly demonstrating the gain of our vectorial representation to model the attention, the use of pairwise combination, and the multi-step iterations in the whole process.  Our final MuRel network is very competitive and outperforms state-of-the-art results on two of the most widely used datasets.


\end{document}